\documentclass[11pt]{article}

\usepackage{ml4or}

\title{
	Machine Learning for Combinatorial Optimization:\\
	a Methodological Tour d'Horizon\thanks{
	    Accepted to the European Journal of Operations Research.
	    \textcopyright 2020. Licensed under the Creative Commons
	    \href{http://creativecommons.org/licenses/by-nc-nd/4.0}{\ccbyncnd}
	}
}
\author[2,3]{Yoshua~Bengio}
\author[1,3]{Andrea~Lodi}
\author[1,3]{Antoine~Prouvost}
\affil[ ]{
    \texttt{yoshua.bengio@mila.quebec} \vskip 0pt
    \texttt{\{andrea.lodi, antoine.prouvost\}@polymtl.ca} \vspace{1em}
}
\affil[1]{Canada Excellence Research Chair in Data Science for Decision Making, École Polytechnique de Montréal}
\affil[2]{Department of Computer Science and Operations Research, Université de Montréal}
\affil[3]{Mila, Quebec Artificial Intelligence Institute}
\date{\vspace{-5ex}}

\begin{document}

\maketitle

\begin{abstract}
	This paper surveys the recent attempts, both from the machine learning and operations research
	communities, at leveraging machine learning to solve combinatorial optimization problems.
	Given the hard nature of these problems, state-of-the-art algorithms rely on handcrafted heuristics
	for making decisions that are otherwise too expensive to compute or mathematically not
	well defined.
	Thus, machine learning looks like a natural candidate to make such decisions in a more principled
	and optimized way.
	We advocate for pushing further the integration of machine learning and combinatorial optimization
	and detail a methodology to do so.
	A main point of the paper is seeing generic optimization problems as data points and inquiring what
	is the relevant distribution of problems to use for learning on a given task.
\end{abstract}

\section{Introduction}
	Operations research, also referred to as prescriptive analytics, started in the second world war as an
	initiative to use mathematics and computer science to assist military planners in their
	decisions \citep{FortunScientists1993}.
	Nowadays, it is widely used in the industry, including but not limited to transportation, supply
	chain, energy, finance, and scheduling.
	In this paper, we focus on discrete optimization problems formulated as integer constrained
	optimization, \textit{i.e.}, with integral or binary variables (called decision variables).
	While not all such problems are hard to solve (\textit{e.g.}, shortest path problems), we
	concentrate on \gls{co} problems (NP-hard).
	This is bad news, in the sense that, for those problems, it is considered unlikely that an algorithm
	whose running time is polynomial in the size of the input exists.
	However, in practice, \gls{co} algorithms can solve instances with up to millions of decision
	variables and constraints.

	\emph{How is it possible to solve NP-hard problems in practical time?}
	Let us look at the example of the \gls{tsp}, a NP-hard problem defined on a graph where we are
	searching for a cycle of minimum length visiting once and only once every node.
	A particular case is that of the \emph{Euclidian} \gls{tsp}.
    In this version, each node is assigned coordinates in a plane,\footnote{
        Or more generally in a vector space of arbitrary dimension.
    } and the cost on an edge connecting two nodes is the Euclidian distance between them.
    While theoretically as hard as the general \gls{tsp}, good approximate solution can be found
    more efficiently in the Euclidian case by leveraging the \emph{structure} of the graph
    \citep[Chapter 6.4.7]{LarsonUrban1981}.
    Likewise, diverse types of problems are solved by leveraging their special structure.
    Other algorithms, designed to be general, are found in hindsight to be empirically more
    efficient on particular problems types.
	The scientific literature covers the rich set of techniques researchers have developed
	to tackle different \gls{co} problems.
	An expert will know how to further refine algorithm parameters to different behaviors
	of the optimization process, thus extending this knowledge with unwritten intuition.
	These techniques, and the parameters controlling them, have been collectively
	\emph{learned} by the community to perform on the inaccessible distribution of problem
	instances deemed valuable.
	The focus of this paper is on \gls{co} algorithms that automatically perform
	learning on a chosen implicit distribution of problems.
	Incorporating \gls{ml} components in the algorithm can achieve this.

	Conversely, \gls{ml} focuses on performing a task given some (finite and usually noisy) data.
	It is well suited for natural signals for which no clear mathematical formulation emerges because
	the true data distribution is not known analytically, such as when processing images, text, voice
	or molecules, or with recommender systems, social networks or financial predictions.
	Most of the times, the learning problem has a statistical formulation that is solved through
	mathematical optimization.
	Recently, dramatic progress has been achieved with deep learning, an \gls{ml} sub-field building
	large parametric approximators by composing simpler functions.
	Deep learning excels when applied in high dimensional spaces with a large number of data points.

	\subsection{Motivation}
		\label{sec:motivation}
		From the \gls{co} point of view, machine learning can help improve an algorithm on a distribution
		of problem instances in two ways.
		On the one side, the researcher assumes expert knowledge\footnote{Theoretical and/or empirical.
		} about the optimization algorithm, but wants to replace some heavy computations by a
		fast approximation.
		Learning can be used to build such approximations in a generic way, \textit{i.e.}, without the need
		to derive new explicit algorithms.
		On the other side, expert knowledge may not be sufficient and some
		algorithmic decisions may be unsatisfactory.
		The goal is therefore to explore the space of these decisions, and learn out of this experience the
		best performing behavior (policy), hopefully improving on the state of the art.
		Even though \gls{ml} is approximate, we will demonstrate through the examples surveyed in this
		paper that this does not systematically mean that incorporating learning will compromise overall
		theoretical guarantees.
		From the point of view of using \gls{ml} to tackle a combinatorial problem, \gls{co} can decompose
		the problem into smaller, hopefully simpler, learning tasks.
		The \gls{co} structure therefore acts as a relevant prior for the model.
		It is also an opportunity to leverage the \gls{co} literature, notably in terms of theoretical
		guarantees (\textit{e.g.}, feasibility and optimality).

	\subsection{Setting}
		Imagine a delivery company in Montreal that needs to solve \glspl{tsp}.
		Every day, the customers may vary, but usually, many are downtown and few on top of the Mont Royal
		mountain.
		Furthermore, Montreal streets are laid on a grid, making the distances close to the $\ell_1$
		distance.
		How close?
		Not as much as Phoenix, but certainly more than Paris.
		The company does not care about solving all possible \glspl{tsp}, but only \emph{theirs}.
		Explicitly defining what makes a \gls{tsp} a likely one for the company is tedious, does not scale, and it
		is not clear how it can be leveraged when explicitly writing an optimization algorithm.
		We would like to automatically specialize \gls{tsp} algorithms for this company.

		The true probability distribution of likely \glspl{tsp} in the Montreal scenario is defining the
		instances on which we would like our algorithm to perform well.
		This is unknown, and cannot even be mathematically characterized in an explicit way.
		Because we do not know what is in this distribution, we can only learn an algorithm that performs
		well on a finite set of \glspl{tsp} sampled from this distribution (for instance, a set of
		historical instances collected by the company), thus implicitly incorporating the desired information about the distribution of instances.
		As a comparison, in traditional \gls{ml} tasks, the true distribution could be that of all possible
		images of cats, while the training distribution is a finite set of such images.
		The challenge in learning is that an algorithm that performs well on problem instances used for
		learning may not work properly on other instances from the true probability distribution.
		For the company, this would mean the algorithm only does well on past problems, but
		not on the future ones.
		To control this, we monitor the performance of the learned algorithm over another independent set
		of \emph{unseen} problem instances.
		Keeping the performances similar between the instances used for learning and the unseen ones is
		known in \gls{ml} as \emph{generalizing}.
		Current \gls{ml} algorithms can generalize to examples from the same distribution, but tend to have
		more difficulty generalizing out-of-distribution (although this is a topic of intense research in ML), and so we may expect \gls{co} algorithms that
		leverage \gls{ml} models to fail when evaluated on unseen problem instances that are too far from
		what has been used for training the \gls{ml} predictor.
		As previously motivated, it is also worth noting that traditional \gls{co} algorithms might not
		even work consistently across all possible instances of a problem family, but rather tend to be
		more adapted to particular structures of problems, \textit{e.g.}, Euclidean \glspl{tsp}.

		Finally, the implicit knowledge extracted by \gls{ml} algorithms is complementary to the hard-won
		explicit expertise extracted through \gls{co} research.
		Rather, it aims to augment and automate the unwritten expert intuition (or lack of) on various
		existing algorithms.
		Given that these problems are highly structured, we believe it is relevant to augment
		solving algorithms with machine learning -- and especially deep learning to address the high
		dimensionality of such problems.

		In the following, we survey the attempts in the literature to achieve such automation and
		augmentation, and we present a methodological overview of those approaches.
		In light of the current state of the field, the literature we survey is exploratory, \textit{i.e.}, we aim at highlighting promising research
		directions in the use of \gls{ml} within \gls{co}, instead of reporting on already mature
		algorithms.

	\subsection{Outline}
		We have introduced the context and motivations for building \acrlong{co} algorithms together with
		\acrlong{ml}.
		The remainder of this paper is organized as follows.
		Section~\ref{sec:preliminaries} provides minimal prerequisites in \acrlong{co}, machine learning,
		deep learning, and reinforcement learning necessary to fully grasp the content of the paper.
		Section~\ref{sec:recent_approaches} surveys the recent literature and derives two distinctive,
		orthogonal, views: Section~\ref{sec:learning_methods} shows how machine learning policies can
		either be learned by imitating an expert or discovered through experience, while Section
		\ref{sec:algorithmic_structure} discusses the interplay between the \gls{ml} and \gls{co}
		components.
		Section~\ref{sec:methodology} pushes further the reflection on the use of \gls{ml} for combinatorial optimization and brings to the fore some methodological points.
		In Section~\ref{sec:challenges}, we detail critical practical challenges of the field.
		Finally, some conclusions are drawn in Section~\ref{sec:conclusions}.

\section{Preliminaries}
	\label{sec:preliminaries}

	In this section, we give a basic (sometimes rough) overview of combinatorial optimization and
	machine learning, with the unique aim of introducing concepts that are strictly required to
	understand the remainder of the paper.

	\subsection{Combinatorial Optimization}
		Without loss of generality, a \gls{co} problem can be formulated as a constrained min-optimization
		program.
		Constraints model natural or imposed restrictions of the problem, variables define the decisions to
		be made, while the objective function, generally a cost to be minimized, defines the measure of the
		quality of every feasible assignment of values to variables.
		If the objective and constraints are linear, the problem is called a \gls{lp} problem.
		If, in addition, some variables are also restricted to only assume integer values, then the problem
		is a \acrfull{milp} problem.

		The set of points that satisfy the constraints is the feasible region.
		Every point in that set (often referred to as a feasible solution) yields an upper bound on the
		objective value of the optimal solution.
		Exact solving is an important aspect of the field, hence a lot of attention is also given to find
		lower bounds to the optimal cost.
		The tighter the lower bounds, with respect to the optimal solution value, the higher the chances
		that the current algorithmic approaches to tackle \glspl{milp} described in the following could be
		successful, \textit{i.e.}, effective if not efficient.

\begin{figure}[H]
			\centering
			\includegraphics[scale=1]{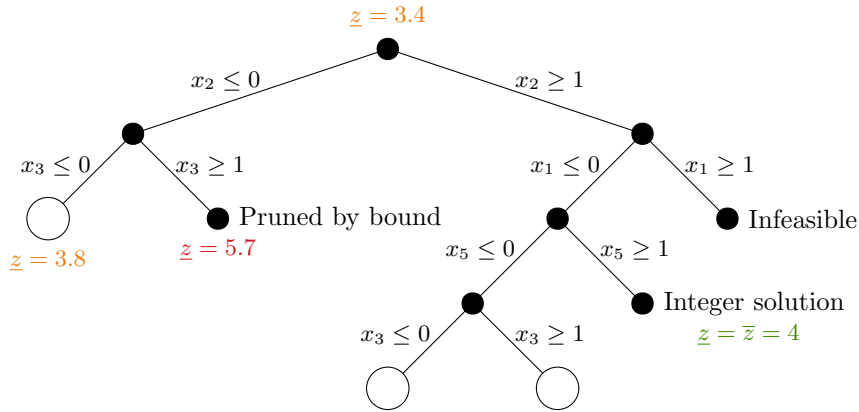}
			\caption{A branch-and-bound tree for \glspl{milp}.
				The \gls{lp} relaxation is computed at every node (only partially shown in the figure).
				Nodes still open for exploration are represented as blank.
			}
			\label{fig:bnb}
		\end{figure}

		Linear and mixed-integer linear programming problems are the workhorse of \gls{co} because they can
		model a wide variety of problems and are the best understood, \textit{i.e.}, there are reliable
		algorithms and software tools to solve them.
		We give them special considerations in this paper but, of course, they do not represent the entire
		\gls{co}, mixed-integer nonlinear programming being a rapidly expanding and very significant area
		both in theory and in practical applications.
		With respect to complexity and solution methods, \gls{lp} is a polynomial problem, well solved, in
		theory and in practice, through the simplex algorithm or interior points methods.
		Mixed-integer linear programming, on the other hand, is an NP-hard problem, which does not make it
		hopeless.
		Indeed, it is easy to see that the complexity of \gls{milp} is associated with the integrality
		requirement on (some of) the variables, which makes the \gls{milp} feasible region nonconvex.
		However, dropping the integrality requirement (i) defines a proper relaxation of \gls{milp}
		(\textit{i.e.}, an optimization problem whose feasible region contains the \gls{milp} feasible
		region), which (ii) happens to be an \gls{lp}, \textit{i.e.}, polynomially solvable.
		This immediately suggests the algorithmic line of attack that is used to solve \gls{milp} through a
		whole ecosystem of \gls{bnb} techniques to perform implicit enumeration.
		Branch and bound implemements a divide-and-conquer type of algorithm representable by a search tree
		in which, at every node, an \gls{lp} relaxation of the problem (possibly augmented by branching
		decisions, see below) is efficiently computed.
		If the relaxation is infeasible, or if the solution of the relaxation is naturally (mixed-)integer,
		\textit{i.e.}, \gls{milp} feasible, the node does not need to be expanded.
		Otherwise, there exists at least one variable, among those supposed to be integer, taking a
		fractional value in the \gls{lp} solution and that variable can be chosen for branching
		(enumeration), \textit{i.e.}, by restricting its value in such a way that two child nodes are
		created.
		The two child nodes have disjoint feasible regions, none of which contains the solution of the
		previous \gls{lp} relaxation.
		We use Figure~\ref{fig:bnb} to illustrate the \gls{bnb} algorithm for a minimization \gls{milp}.
		At the root node in the figure, the variable $x_2$ has a fractional value in the \gls{lp} solution
		(not represented), thus branching is done on the floor (here zero) and ceiling (here one) of this
		value.
		When an integer solution is found, we also get an upper bound (denoted as $\overline{z}$) on the
		optimal solution value of the problem.
		At every node, we can then compare the solution value of the relaxation (denoted as
		$\underline{z}$) with the minimum upper bound found so far, called the incumbent solution value.
		If the latter is smaller than the former for a specific node, no better (mixed-)integer solution
		can be found in the sub-tree originated by the node itself, and it can be pruned.

		All commercial and noncommercial \gls{milp} solvers enhance the above enumeration framework with
		the extensive use of cutting planes, \emph{i.e.}, valid linear inequalities that are added to the
		original formulation (especially at the root of the \gls{bnb} tree) in the attempt of strengthening
		its \gls{lp} relaxation.
		The resulting framework, referred to as the branch-and-cut algorithm, is then further enhanced by
		additional algorithmic components, preprocessing and primal heuristics being the most crucial ones.
		The reader is referred to \cite{wolsey1998integer} and \cite{Conforti2014} for extensive textbooks
		on \gls{milp} and to \cite{Lodi2008} for a detailed description of the algorithmic components of
		the \gls{milp} solvers.

		We end the section by noting that there is a vast literature devoted to (primal) heuristics,
		\textit{i.e.}, algorithms designed to compute ``good in practice" solutions to CO problems
		without optimality guarantee.
		Although a general discussion on them is outside the scope here, those heuristic methods play
		a central role in \gls{co} and will be considered in specific contexts in the present paper.
		The interested reader is referred to \cite{FischettiLodi2011} and \cite{gendreau2010handbook}.

	\subsection{Machine Learning}
		\label{sec:maching_learning}
		\paragraph{Supervised learning}
		In supervised learning, a set of input (features) / target pairs is provided and the task is to
		find a function that for every input has a predicted output as close as possible to the provided
		target.
		Finding such a function is called learning and is solved through an optimization problem over a
		family of functions.
		The loss function, \textit{i.e.}, the measure of discrepancy between the output and the target, can
		be chosen depending on the task (regression, classification, \textit{etc.}) and on the optimization
		methods.
		However, this approach is not enough because the problem has a statistical nature.
		It is usually easy enough to achieve a good score on the given examples but one wants to achieve a
		good score on unseen examples (test data).
		This is known as generalization.

		Mathematically speaking, let $X$ and $Y$, following a joint probability distribution $P$, be random
		variables representing the input features and the target.
		Let $\ell$ be the per sample loss function to minimize, and let $\{f_\theta \, | \, \theta \in
			\mathbb{R}^p\}$ be the family of \gls{ml} models (parametric in this case) to optimize over.
		The supervised learning problem is framed as
		\begin{equation}
			\label{eq:supervised}
			\min_{\theta \in \mathbb{R}^p}\ \mathbb{E}_{X, Y \sim P}\, \ell(Y, f_\theta(X)).
		\end{equation}
		For instance,
		$f_\theta$ could be a linear model with weights $\theta$ that we wish to \emph{learn}.
		The loss function $\ell$ is task dependent (\textit{e.g.}, classification error) and can sometimes be
		replaced by a surrogate one (\textit{e.g.}, a differentiable one).
		The probability distribution is unknown and inaccessible.
		For example, it can be the probability distribution of all natural images.
		Therefore, it is approximated by the empirical probability distribution over a finite dataset
		$D_{train} = \{(x_i, y_i)\}_i$ and the optimization problem solved is
		\begin{equation}
			\label{eq:supervised_empirical}
			\min_{\theta \in \mathbb{R}^p}\ \sum_{(x, y) \in D_{train}}
			\frac{1}{|D_{train}|} \ell(y, f_\theta(x)).
		\end{equation}
		A model is said to generalize, if low
		objective values of \eqref{eq:supervised_empirical} translate in low objective values of
	    \eqref{eq:supervised}.
		Because \eqref{eq:supervised} remains inaccessible, we estimate the generalization error by
		evaluating the trained model on a separate test dataset $D_{test}$ with
		\begin{equation}
			\sum_{(x, y) \in D_{test}} \frac{1}{|D_{test}|} \ell(y, f_\theta(x)).
		\end{equation}
		If a model (\textit{i.e.}, a
		family of functions) can represent many different functions, the model is said to have high
		capacity and is prone to overfitting: doing well on the training data but not generalizing to the
		test data.
		Regularization is anything that can improve the test score at the expense of the training score and
		is used to restrict the practical capacity of a model.
		On the contrary, if the capacity is too low, the model underfits and performs poorly on both sets.
		The boundary between overfitting and underfitting can be estimated by changing the effective
		capacity (the richness of the family of functions reachable by training): below the critical
		capacity one underfits and test error decreases with increases in capacity, while above that
		critical capacity one overfits and test error increases with increases in capacity.

		Selecting the best among various trained models cannot be done on the test set.
		Selection is a form of optimization, and doing so on the test set would bias the estimator in
		\eqref{eq:supervised_empirical}.
		This is a common form of data dredging, and a mistake to be avoided.
		To perform model selection, a validation dataset $D_{valid}$ is used to estimate the generalization
		error of different \gls{ml} models is necessary.
		Model selection can be done based on these estimates, and the final unbiased generalization error
		of the selected model can be computed on the test set. The validation set is therefore often used
		to select effective capacity, \textit{e.g.}, by changing the amount of training, the number of parameters
		$\theta$, and the amount of regularization imposed to the model.

		\paragraph{Unsupervised learning}
		In unsupervised learning, one does not have targets for the task one wants to solve, but rather
		tries to capture some characteristics of the joint distribution of the observed random variables.
		The variety of tasks include density estimation, dimensionality reduction, and clustering.
		Because unsupervised learning has received so far little attention in conjunction with \gls{co} and
		its immediate use seems difficult, we are not discussing it any further.
		The reader is referred to \cite{bishop2006pattern,murphy2012machine,goodfellow2016deep} for
		textbooks on machine learning.

\begin{figure}[H]
			\centering
			\includegraphics[scale=.8]{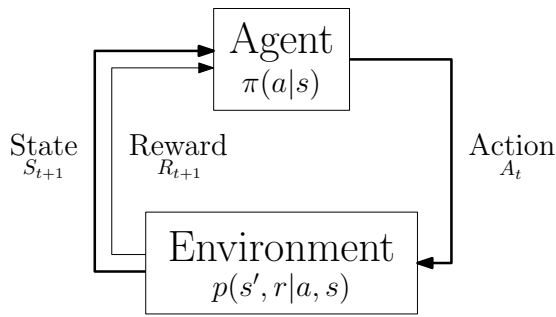}
			\caption{The Markov decision process associated with reinforcement learning,
				modified from \cite{sutton2018reinforcement}.
				The agent behavior is defined by its policy $\pi$, while the environment evolution is defined by
				the dynamics $p$.
				Note that the reward is not necessary to define the evolution and is provided only as a learning
				mechanism for the agent.
				Actions, states, and rewards are random variables in the general framework.
			}
			\label{fig:mdp}
		\end{figure}

		\paragraph{Reinforcement learning}
		In \gls{rl}, an agent interacts with an environment through a \gls{mdp}, as illustrated in Figure
		\ref{fig:mdp}.
		At every time step, the agent is in a given state of the environment and chooses an action
		according to its (possibly stochastic) policy.
		As a result, it receives from the environment a reward and enters a new state.
		The goal in \gls{rl} is to train the agent to maximize the expected sum of future rewards, called the
		return.
		For a given policy, the expected return given a current state (resp.\@ state and action pair) is
		known as the value function (resp.\@ state action value function).
		Value functions follow the Bellman equation, hence the problem can be formulated as dynamic
		programming, and solved approximately.
		The dynamics of the environment need not be known by the agent and are learned directly or
		indirectly, yielding an exploration \textit{vs} exploitation dilemma: choosing between exploring
		new states for refining the knowledge of the environment for possible long-term improvements, or
		exploiting the best-known scenario learned so far (which tends to be in already visited or
		predictable states).

		The state should fully characterize the environment at every step, in the sense that future
		states only depend on past states via the current state (the Markov property).
		When this is not the case, similar methods can be applied but we say that the agent receives an
		\emph{observation} of the state.
		The Markov property no longer holds and the \gls{mdp} is said to be partially observable.

		Defining a reward function is not always easy.
		Sometimes one would like to define a very sparse reward, such as 1 when the agent solves the
		problem, and 0 otherwise.
		Because of its underlying dynamic programming process, \gls{rl} is naturally able to credit
		states/actions that lead to future rewards.
		Nonetheless, the aforementioned setting is challenging as it provides no learning opportunity until
		the agent (randomly, or through advanced approaches) solves the problem.
		Furthermore, when the policy is approximated (for instance, by a linear function), the learning is
		not guaranteed to converge and may fall into local minima.
		For example, an autonomous car may decide not to drive anywhere for fear of hitting a pedestrian
		and receiving a dramatic negative reward.
		These challenges are strongly related to the aforementioned exploration dilemma.
		The reader is referred to \cite{sutton2018reinforcement} for an extensive textbook on reinforcement
		learning.

		\paragraph{Deep learning}
		Deep learning is a successful method for building parametric composable functions in high
		dimensional spaces.
		In the case of the simplest neural network architecture, the feedforward neural network
		(also called an \gls{mlp}), the input data is successively passed through a number of layers.
		For every layer, an affine transformation is applied on the input vector, followed by a non-linear
		scalar function (named activation function) applied element-wise.
		The output of a layer, called intermediate activations, is passed on to the next layer.
		All affine transformations are independent and represented in practice as different matrices of
		coefficients.
		They are learned, \textit{i.e.}, optimized over, through \gls{sgd}, the optimization algorithm used
		to minimize the selected loss function.
		The stochasticity comes from the limited number of data points used to compute the loss before
		applying a gradient update.
		In practice, gradients are computed using reverse mode automatic differentiation, a practical
		algorithm based on the chain rule, also known as back-propagation.
		Deep neural networks can be difficult to optimize, and a large variety of techniques have been
		developed to make the optimization behave better, often by changing architectural designs of the
		network.
		Because neural networks have dramatic capacities, \textit{i.e.}, they can essentially match any
		dataset, thus being prone to overfitting, they are also heavily regularized.
		Training them by \gls{sgd} also regularizes them because of the noise in the gradient, making
		neural networks generally robust to overfitting issues, even when they are very large and would
		overfit if trained with more aggressive optimization.
		In addition, many hyper-parameters exist and different combinations are evaluated (known as
		hyper-parameters optimization).
		Deep learning also sets itself apart from more traditional \gls{ml} techniques by taking as inputs
		all available raw features of the data, \textit{e.g.}, all pixels of an image, while traditional
		\gls{ml} typically requires to engineer a limited number of domain-specific features.

\begin{figure}[H]
			\centering
			\begin{subfigure}{.4\textwidth}
				\centering
				\includegraphics[scale=.5]{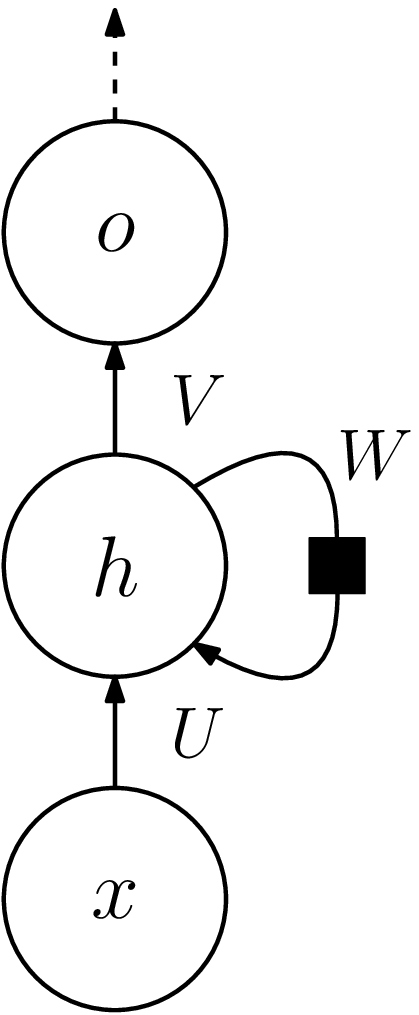}
			\end{subfigure}
			\begin{subfigure}{.4\textwidth}
				\centering
				\includegraphics[scale=.5]{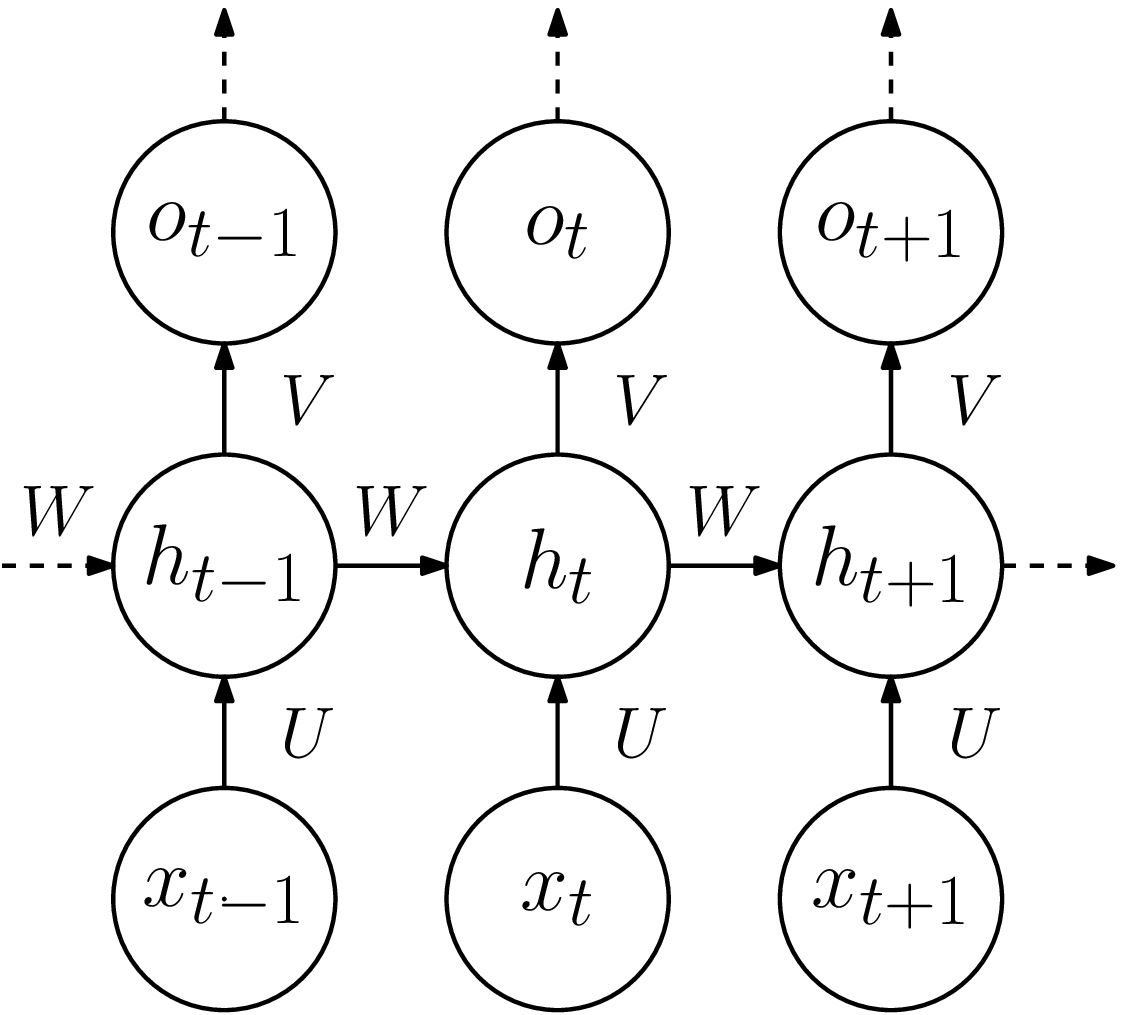}
			\end{subfigure}
			\caption{A vanilla \gls{rnn} modified from \cite{goodfellow2016deep}.
				On the left, the black square indicates a one step delay.
				On the right, the same \gls{rnn} is shown unfolded.
				Three sets $U$, $V$, and $W$ of parameters are represented and re-used at every time step.
			}
			\label{fig:rnn}
		\end{figure}

		Deep learning researchers have developed different techniques to tackle this variety of structured
		data in a manner that can handle variable-size data structures, \textit{e.g.},
		variable-length sequences.
		In this paragraph, and in the next, we present such state-of-the-art techniques.
		These are complex topics, but lack of comprehension does not hinder the reading of the paper.
		At a high level, it is enough to comprehend that these are architectures designed to handle different
		structures of data.
		Their usage, and in particular the way they are learned, remains very similar to plain
		feedforward neural networks introduced above.
		The first architectures presented are the \glspl{rnn}.
		These models can operate on sequence data by \emph{sharing parameters} across different sequence
		steps.
		More precisely, a same neural network block is successively applied at every step of the sequence,
		\textit{i.e.}, with the same architecture and parameter values at each time step.
		The specificity of such a network is the presence of recurrent layers: layers that take as input
		both the activation vector of the previous layer and its own activation vector on the preceding
		sequence step (called a hidden state vector), as illustrated in Figure~\ref{fig:rnn}.

\begin{figure}[H]
			\centering
			\includegraphics[scale=.8]{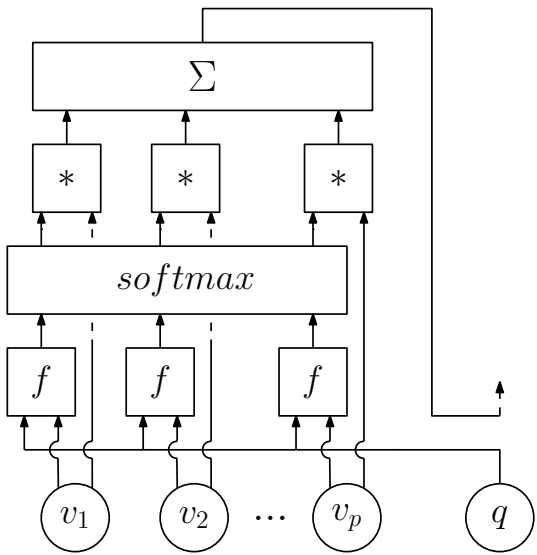}
			\caption{
				A vanilla attention mechanism where a query $q$ is computed against a set of values
				$(v_i)_i$.
				An affinity function $f$, such as a dot product, is used on query and value pairs.
				If it includes some parameters, the mechanism can be learned.
			}
			\label{fig:attention}
		\end{figure}

		Another important size-invariant technique are \emph{attention mechanisms}.
		They can be used to process data where each data point corresponds to a set.
		In that context, parameter sharing is used to address the fact that different sets need not to be
		of the same size.
		Attention is used to query information about all elements in the set, and merge it for downstream
		processing in the neural
		network, as depicted in Figure~\ref{fig:attention}.
		An affinity function takes as input the query (which represents any kind of contextual information
		which informs where attention should be concentrated) and a representation of an element of the set
		(both are activation vectors) and outputs a scalar.
		This is repeated over all elements in the set for the same query.
		Those scalars are normalized (for instance with a softmax function) and used to define a weighted
		sum of the representations of elements in the set that can, in turn, be used in the neural network
		making the query. This form of content-based soft attention was introduced
		by~\citet{Bahdanau-et-al-ICLR2015-small}.
		A general explanation of attention mechanisms is given by \cite{VaswaniAttentionAllyou2017}.
		Attention can be used to build \glspl{gnn}, \textit{i.e.}, neural networks able to process graph
		structured input data, as done by \cite{VelickovicGraphAttention2018}.
		In this architecture, every node attends over the set of its neighbors.
		The process is repeated multiple times to gather information about nodes further away.
		\Glspl{gnn} can also be understood as a form of message passing
		\citep{GilmerNeuralMessagePassing2017}.

		Deep learning and back-propagation can be used in supervised, unsupervised, or reinforcement learning.
		The reader is referred to \cite{goodfellow2016deep} for a machine learning textbook
		devoted to deep learning.

\section{Recent approaches}
	\label{sec:recent_approaches}
	We survey different uses of \gls{ml} to help solve combinatorial optimization problems and organize
	them along two orthogonal axes.
	First, in Section~\ref{sec:learning_methods} we illustrate the two main motivations for using
	learning: approximation and discovery of new policies.
	Then, in Section~\ref{sec:algorithmic_structure}, we show examples of different ways to combine
	learned and traditional algorithmic elements.

	\subsection{Learning methods}
		\label{sec:learning_methods}
		This section relates to the two motivations reported in Section~\ref{sec:motivation} for using
		\gls{ml} in \gls{co}.
		In some works, the researcher assumes theoretical and/or empirical knowledge about the decisions to
		be made for a \gls{co} algorithm, but wants to alleviate the computational burden by approximating
		some of those decisions with machine learning.
		On the contrary, we are also motivated by the fact that, sometimes, expert knowledge is not
		satisfactory and the researcher wishes to find better ways of making decisions.
		Thus, \gls{ml} can come into play to train a model through trial and error reinforcement learning.

		We frame both these motivations in the state/action \gls{mdp} framework introduced in section
		\ref{sec:maching_learning}, where the environment is the internal state of the algorithm.
		We care about learning algorithmic decisions utilized by a \gls{co} algorithm and we call the
		function making the decision a \textit{policy}, that, given all available information,\footnote{
		    A \textit{state} if the information is sufficient to fully characterize the environment at that
		    time in a Markov decision process setting.
		} returns (possibly stochastically) the action to be taken.
		The policy is the function that we want to learn using \gls{ml} and we show in the following how
		the two motivations naturally yield two learning settings.
		Note that the case where the length of the trajectory of the \gls{mdp} has value 1 is a common edge
		case (called the bandit setting) where this formulation can seem excessive, but it nonetheless helps
		comparing methods.

\begin{figure}[H]
			\centering
			\includegraphics[scale=.7]{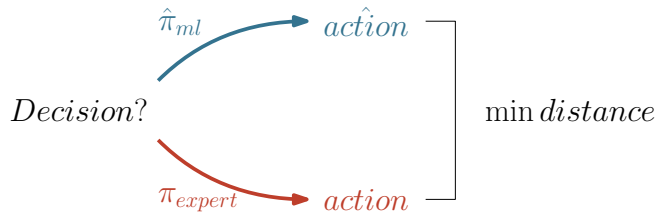}
			\caption{In the demonstration setting, the policy is trained to reproduce the
				action of an expert policy by minimizing some discrepancy in the action space.}
			\label{fig:demonstration}
		\end{figure}

		In the case of using \gls{ml} to approximate decisions, the policy is often learned by
		\emph{imitation learning}, thanks to \emph{demonstrations}, because the expected behavior is shown
		(demonstrated) to the \gls{ml} model by an expert (also called oracle, even though it is not
		necessarily optimal), as shown in Figure~\ref{fig:demonstration}.
		In this setting, the learner is not trained to optimize a performance measure, but to
		\emph{blindly} mimic the expert.

\begin{figure}[H]
			\centering
			\includegraphics[scale=.7]{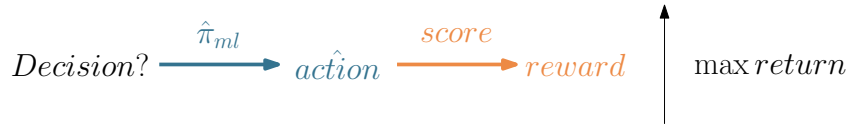}
			\caption{When learning through a reward signal, no expert is involved; only
				maximizing the expected sum of future rewards (the return) matters.}
			\label{fig:experience}
		\end{figure}

		In the case where one cares about discovering new policies, \textit{i.e.}, optimizing an
		algorithmic decision function from the ground up, the policy may be learned by \acrlong{rl} through
		\textit{experience}, as shown in Figure~\ref{fig:experience}.
		Even though we present the learning problem under the fundamental \gls{mdp} of \gls{rl}, this does
		not constrain one to use the major \gls{rl} algorithms (approximate dynamic programming and
		policy gradients) to maximize the expected sum of rewards.
		Alternative optimization methods, such as bandit algorithms, genetic algorithms, direct/local
		search, can also be used to solve the \gls{rl} problem.\footnote{
			In general, identifying which algorithm will perform best is an open research question
			unlikely to have a simple answer, and is outside of the scope of the methodology presented here.
		}

		It is critical to understand that in the imitation setting, the policy is learned through
		supervised targets provided by an expert for every action (and without a reward), whereas in the
		experience setting, the policy is learned from a reward (possibly delayed) signal using \gls{rl}
		(and without an expert).
		In imitation, the agent is taught \emph{what} to do, whereas in \gls{rl}, the agent is encouraged
		to quickly \emph{accumulate} rewards.
		The distinction between these two settings is far more complex than the distinction made here.
		We explore some of this complexity, including their strengths and weaknesses, in Section
		\ref{sec:demonstration_and_experience}.

		In the following, few papers demonstrating the different settings are surveyed.

		\subsubsection{Demonstration}
		\label{sec:demonstration}
		In  \cite{Baltean-LugojanStrongsparsecut2018}, the authors use a neural network to approximate the
		lower bound improvement generated by tightening the current relaxation via cutting planes (cuts,
		for short).
		More precisely, \cite{Baltean-LugojanStrongsparsecut2018} consider non-convex quadratic programming
		problems and aim at approximating the associated \gls{sdp} relaxation, known to be strong but
		time-consuming, by a linear program.
		A straightforward way of doing that is to iteratively add (linear) cutting planes associated with
		negative eigenvalues, especially considering small-size (square) submatrices of the original
		quadratic objective function.
		That approach has the advantage of generating sparse cuts\footnote{
		    The reader is referred to \cite{DeyMolinaro2018} for a detailed discussion on the importance of
		    sparse cutting planes in \gls{milp}.
		} but it is computationally challenging because of the exponential number of those
		submatrices and because of the difficulty of finding the right metrics to select among the violated
		cuts.
		The authors propose to solve \glspl{sdp} to compute the bound improvement associated with
		considering specific submatrices, which is also a proxy on the quality of the cuts that could be
		separated from the same submatrices.
		In this context, supervised (imitation) learning is applied offline to approximate the objective
		value of the \gls{sdp} problem associated with a submatrix selection and, afterward, the model can
		be rapidly applied to select the most promising submatrices without the very significant
		computational burden of solving \glspl{sdp}.
		Of course, the rational is that the most promising submatrices correspond to the most promising
		cutting planes and \cite{Baltean-LugojanStrongsparsecut2018} train a model to estimate the
		objective of an \gls{sdp} problem only in order to decide to add the most promising cutting planes.
		Hence, cutting plane selection is the ultimate policy learned.

		Another example of demonstration is found in the context of branching policies in \gls{bnb} trees
		of \glspl{milp}.
		The choice of variables to branch on can dramatically change the size of the \gls{bnb} tree, hence
		the solving time.
		Among many heuristics, a well-performing approach is \emph{strong branching} \citep{applegate2007}.
		Namely, for every branching decision to be made, strong branching performs a one step look-ahead by
		tentatively branching on many candidate variables, computes the \gls{lp} relaxations to get the
		potential lower bound improvements, and eventually branches on the variable providing the best
		improvement.
		Even if not all variables are explored, and the \gls{lp} value can be approximated, this is still a
		computationally expensive strategy.
		For these reasons, \cite{MarcosAlvarezsupervisedmachinelearning2014,
		MarcosAlvarezMachineLearningBasedApproximation2017} use a special type of decision tree (a
		classical model in supervised learning) to approximate strong branching decisions using supervised
		learning.
		\cite{KhalilLearningbranchmixed2016} propose a similar approach, where a linear model is
		learned on the fly for every instance by using strong branching at the top of the tree, and
		eventually replacing it by its \gls{ml} approximation.
		The linear approximator of strong branching introduced in
		\cite{MarcosAlvarezOnlineLearningStrong2016} is learned in an active fashion: when the estimator is
		deemed unreliable, the algorithm falls back to true strong branching and the results are then used
		for both branching and learning.
		In all the branching algorithms presented here, inputs to the \gls{ml} model are engineered as a
		vector of fixed length with static features descriptive of the instance, and dynamic features
		providing information about the state of the \gls{bnb} process.
		\cite{GasseExact2019} use a neural network to learn an offline approximation to strong
		branching, but, contrary to the aforementioned papers, the authors use a raw exhaustive
		representation (\textit{i.e.}, they do not discard nor aggregate any information) of the
		sub-problem associated with the current branching node as input to the \gls{ml} model.
		Namely, an \gls{milp} sub-problem is represented as a bipartite graph on variables and constraints,
		with edges representing non-zero coefficients in the constraint matrix.
		Each node is augmented with a set of features to fully describe the sub-problem, and a \gls{gnn} is
		used to build an \gls{ml} approximator able to process this type of structured data.
		Node selection, \textit{i.e.}, deciding on the next branching node to explore in a \gls{bnb} tree,
		is also a critical decision in \gls{milp}.
		\cite{HeLearningSearchBranch2014} learn a policy to select among the open branching
		nodes the one that contains the optimal solution in its sub-tree.
		The training algorithm is an online learning method collecting expert behaviors throughout the
		entire learning phase.
		The reader is referred to \cite{Lodilearningbranchingsurvey2017} for an extended survey on learning
		and branching in \glspl{milp}.

		Branch and bound is a technique not limited to \gls{milp} and can be use for general
		tree search.
		\cite{HottungDeep2017} build a tree search procedure for the container pre-marshalling
		problem in which they aim to learn, not only a branching policy (similar in principle to what has
		been discussed in the previous paragraph), but also a value network to estimate the value of
		partial solutions and used for bounding.
		The authors leverage a form of \gls{cnn}\footnote{
		    A type of neural network, usually used on image input, that leverages parameter sharing to
		    extract local information.
		} for both networks and train them in a supervised fashion using pre-computed solutions
		of the problem.
		The resulting algorithm is heuristic due the approximations made while bounding.

		As already mentioned at the beginning of Section~\ref{sec:learning_methods}, learning a policy by
		demonstration is identical to supervised learning, where training pairs of input state and target
		actions are provided by the expert.
		In the simplest case, expert decisions are collected beforehand, but more advanced methods can
		collect them online to increase stability as previously shown in
		\cite{MarcosAlvarezOnlineLearningStrong2016} and \cite{HeLearningSearchBranch2014}.

		\subsubsection{Experience}
		\label{sec:experience}
		Considering the \gls{tsp} on a graph, it is easy to devise a greedy heuristic that builds a tour by
		sequentially picking the nodes among those that have not been visited yet, hence defining a
		permutation.
		If the criterion for selecting the next node is to take the closest one, then the heuristic is
		known as the nearest neighbor.
		This simple heuristic has poor practical performance and many other heuristics perform better
		empirically, \textit{i.e.}, build cheaper tours.
		Selecting the nearest node is a fair intuition but turns out to be far from satisfactory.
		\cite{KhalilLearningCombinatorialOptimization2017} suggest learning the criterion for this
		selection.
		They build a greedy heuristic framework, where the node selection policy is learned using a
		\gls{gnn} \citep{DaiDiscriminativeEmbeddingsLatent2016}, a type of neural network able to process
		input graphs of any finite size by a mechanism of message passing \citep{GilmerNeuralMessagePassing2017}.
		For every node to select, the authors feed to the network the graph representation of the problem
		-- augmented with features indicating which of the nodes have already been visited -- and receive
		back an action value for every node.
		Action values are used to train the network through \gls{rl} (Q-learning in particular) and the
		partial tour length is used as a reward.

		This example does not do justice to the rich \gls{tsp} literature that has developed far more
		advanced algorithms performing orders of magnitude better than \gls{ml} ones.
		Nevertheless, the point we are trying to highlight here is that given a fixed context, and a
		decision to be made, \gls{ml} can be used to discover new, potentially better performing policies.
		Even on state-of-the-art \gls{tsp} algorithms (\textit{i.e.}, when exact solving is taken to its
		limits), many decisions are made in heuristic ways, \textit{e.g.}, cutting plane selection, thus
		leaving room for \gls{ml} to assist in making these decisions.

		Once again, we stress that learning a policy by experience is well described by the \gls{mdp}
		framework of reinforcement learning, where an agent maximizes the return (defined in Section
		\ref{sec:maching_learning}).
		By matching the reward signal with the optimization objective, the goal of the learning agent
		becomes to solve the problem, without assuming any expert knowledge.
		Some methods that were not presented as \gls{rl} can also be cast in this \gls{mdp}
		formulation, even if the optimization methods are not those of the \gls{rl} community.
		For instance, part of the \gls{co} literature is dedicated to automatically build specialized
		heuristics for different problems.
		The heuristics are build by orchestrating a set of moves, or subroutines, from a pre-defined
		domain-specific collections.
		For instance, to tackle bipartite boolean quadratic programming problems,
		\cite{KarapetyanMarkov2017} represent this orchestration as a Markov chain where the states are the
		subroutines.
		One Markov chain is parametrized by its transition probabilities.
		\cite{MasciaGrammar2014}, on the other hand, define valid succession of moves
		through a grammar, where words are moves and sentences correspond to heuristics.
		The authors introduce a parametric space to represent sentences of a grammar.
		In both cases, the setting is very close to the \gls{mdp} of \gls{rl}, but the parameters are
		learned though direct optimization of the performances of their associated heuristic through
		so-called \emph{automatic configuration tools} (usually based on genetic or local search, and
		exploiting parallel computing).
		Note that the learning setting is rather simple as the parameters do not adapt to the problem
		instance, but are fixed for various clusters.
		From the \gls{ml} point of view, this is equivalent to a piece-wise constant regression.
		If more complex models were to be used, direct optimization may not scale adequately to obtain good
		performances.
		The same approach to building heuristics can be brought one level up if, instead of orchestrating
		sets of moves, it arranges predefined heuristics.
		The resulting heuristic is then called a \textit{hyper-heuristic}.
		\cite{OzcanReinforcement2012} build a hyper-heuristic for examination timetabling
		by learning to combine existing heuristics.
		They use a bandit algorithm, a stateless form of \gls{rl}
		\citep[see][Chapter~2]{sutton2018reinforcement}, to learn online a value function for each
		heuristic.
		\\

		We close this section by noting that demonstration and experience are not mutually exclusive and
		most learning tasks can be tackled in both ways.
		In the case of selecting the branching variables in an \gls{milp} branch-and-bound tree, one could
		adopt anyone of the two prior strategies.
		On the one hand, \cite{MarcosAlvarezsupervisedmachinelearning2014,
		MarcosAlvarezOnlineLearningStrong2016, MarcosAlvarezMachineLearningBasedApproximation2017,
		KhalilLearningbranchmixed2016} estimate that strong branching is an effective branching strategy
		but computationally too expensive and build a machine learning model to approximate it.
		On the other hand, one could believe that no branching strategy is good enough and try to learn one
		from the ground up, for instance through reinforcement learning as suggested (but not implemented)
		in \cite{KhalilLearningbranchmixed2016}.
		An intermediary approach is proposed by \cite{LibertoDash2016}.
		The authors recognize that, among traditional variable selection policies, the ones performing well
		at the top of the \gls{bnb} tree are not necessarily the same as the ones performing well deeper
		down.
		Hence, the authors learn a model to dynamically switch among predefined policies during \gls{bnb}
		based on the current state of the tree.
		While this seems like a case of imitation learning, given that traditional branching policies can
		be thought of as experts, this is actually not the case.
		In fact, the model is not learning \emph{from} any expert, but really learning to choose between
		pre-existing policies.
		This is technically not a branching variable selection, but rather a branching heuristic selection
		policy.
		Each sub-tree is represented by a vector of handcrafted features, and a clustering of these vectors
		is performed.
		Similarly to what was detailed in the previous paragraph about the work of
		\cite{KarapetyanMarkov2017, MasciaGrammar2014}, automatic configuration tools are then used to
		assign the best branching policy to each cluster.
		When branching at a given node, the cluster the closest to the current sub-tree is retrieved,
		and its assigned policy is used.

	\subsection{Algorithmic structure}
		\label{sec:algorithmic_structure}
		In this section, we survey how the learned policies (whether from demonstration or experience) are
		combined with traditional \gls{co} algorithms, \textit{i.e.}, considering \gls{ml} and explicit
		algorithms as building blocks, we survey how they can be laid out in different templates.
		The three following sections are not necessarily disjoint nor exhaustive but are a natural way to
		look at the literature.

		\subsubsection{End to end learning}
		\label{sec:end2end_learning}
\begin{figure}[H]
			\centering
			\includegraphics[scale=.85]{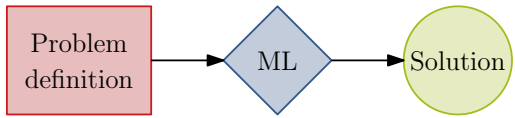}
			\caption{Machine learning acts alone to provide a solution to the problem.}
			\label{fig:ml_alone}
		\end{figure}

		A first idea to leverage machine learning to solve discrete optimization problems is to train the
		\gls{ml} model to output solutions directly from the input instance, as shown in Figure
		\ref{fig:ml_alone}.

		This approach has been explored recently, especially on Euclidean \glspl{tsp}.
		To tackle the problem with deep learning, \cite{VinyalsPointerNetworks2015} introduce the pointer
		network wherein an encoder, namely an \gls{rnn}, is used to parse all the \gls{tsp} nodes in the
		input graph and produces an encoding (a vector of activations) for each of them.
		Afterward, a decoder, also an \gls{rnn}, uses an attention mechanism similar
		to~\citet{Bahdanau-et-al-ICLR2015-small} (Section~\ref{sec:maching_learning}) over the previously
		encoded nodes in the graph to produce a probability
		distribution over these nodes (through the softmax layer previously illustrated in Figure
		\ref{fig:attention}).
		Repeating this decoding step, it is possible for the network to output a permutation over its
		inputs (the \gls{tsp} nodes).
		This method makes it possible to use the network over different input graph sizes.
		The authors train the model through supervised learning with precomputed \gls{tsp} solutions as
		targets.
		\cite{BelloNeuralCombinatorialOptimization2016} use a similar model and train it with
		reinforcement learning using the tour length as a reward signal.
		They address some limitations of supervised (imitation) learning, such as the need to compute
		optimal (or at least high quality) \gls{tsp} solutions (the targets), that in turn, may be
		ill-defined when those solutions are not computed exactly, or when multiple solutions exist.
		\cite{KoolAttentionSolvesYour2018} introduce more prior knowledge in the model using a \gls{gnn}
		instead of an \gls{rnn} to process the input.
		\cite{EmamiLearningPermutationsSinkhorn2018} and \cite{NowakNoteLearningAlgorithms2017}
		explore a different approach by directly approximating a double stochastic matrix in the output of
		the neural network to characterize the permutation.
		The work of \cite{KhalilLearningCombinatorialOptimization2017}, introduced in Section
		\ref{sec:experience}, can also be understood as an end to end method to tackle the \gls{tsp}, but
		we prefer to see it under the eye of Section~\ref{sec:learning_alongside}.
		It is worth noting that tackling the \gls{tsp} through \gls{ml} is not new.
		Earlier work from the nineties focused on Hopfield neural networks and self organizing neural
		networks, the interested reader is referred to the survey of
		\cite{SmithNeuralNetworksCombinatorial1999}.

		In another example, \cite{LarsenPredictingSolutionSummaries2018} train a neural network to predict
		the solution of a stochastic load planning problem for which a deterministic \gls{milp} formulation
		exists.
		Their main motivation is that the application needs to make decisions at a tactical level,
		\textit{i.e.}, under incomplete information, and machine learning is used to address the
		stochasticity of the problem arising from missing some of the state variables in the observed input.
		The authors use operational solutions, \textit{i.e.}, solutions to the deterministic version of the
		problem, and aggregate them to provide (tactical) solution targets to the \gls{ml} model.
		As explained in their paper, the highest level of description of the solution is its cost, whereas
		the lowest (operational) is the knowledge of values for all its variables.
		Then, the authors place themselves in the middle and predict an aggregation of variables (tactical)
		that corresponds to the stochastic version of their specific problem.
		Furthermore, the nature of the application requires to output solutions in real time, which is not
		possible either for the stochastic version of the load planning problem or its deterministic
		variant when using state-of-the-art \gls{milp} solvers.
		Then, \gls{ml} turns out to be suitable for obtaining accurate solutions with short computing times
		because some of the complexity is addressed offline, \textit{i.e.}, in the learning phase, and the
		run-time (inference) phase is extremely quick.
		Finally, note that in \cite{LarsenPredictingSolutionSummaries2018} an \acrshort{mlp},
		\textit{i.e.}, a feedforward neural network, is used to process the input instance as a
		vector, hence integrating very little prior knowledge about the problem structure.

		\subsubsection{Learning to configure algorithms}
		\label{sec:algorithm_configuration}
\begin{figure}[H]
			\centering
			\includegraphics[scale=.85]{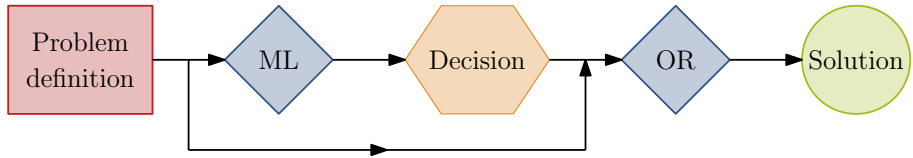}
			\caption{The machine learning model is used to augment an operation research
				algorithm with valuable pieces of information.}
			\label{fig:ml_then_or}
		\end{figure}

		In many cases, using only machine learning to tackle the problem may not be the most suitable
		approach.
		Instead, \gls{ml} can be applied to provide additional pieces of information to a \gls{co}
		algorithm as illustrated in Figure~\ref{fig:ml_then_or}.
		For example, \gls{ml} can provide a parametrization of the algorithm (in a very broad sense).

		Algorithm configuration, detailed in \cite{HoosAutomated2012, BischlAslib2016}, is a well studied area that captures the setting presented here.
		Complex optimization algorithms usually have a set of parameters left constant during optimization
		(in machine learning they are called hyper-parameters).
		For instance, this can be the aggressiveness of the pre-solving operations (usually controlled by a
		single parameter) of an \gls{milp} solver, or the learning rate / step size in gradient descent
		methods.
		Carefully selecting their value can dramatically change the performance of the optimization
		algorithm.
		Hence, the algorithm configuration community started looking for good default parameters.
		Then good default parameters for different cluster of similar problem instances.
		From the \gls{ml} point of view, the former is a constant regression, while the second is a
		piece-wise constant nearest neighbors regression.
		The natural continuation was to learn a regression mapping problem instances to algorithm
		parameters.

		In this context, \cite{KruberLearningWhenUse2017} use machine learning on \gls{milp} instances to
		estimate beforehand whether or not applying a Dantzig-Wolf decomposition will be effective,
		\textit{i.e.}, will make the solving time faster.
		Decomposition methods can be powerful but deciding if and how to apply them depends on many
		ingredients of the instance and of its formulation and there is no clear cut way of optimally
		making such a decision.
		In their work, a data point is represented as a fixed length vector with features representing
		instance and tentative decomposition statistics.
		In another example, in the context of \acrlong{miqp},
		\cite{BonamiLearningClassificationMixedInteger2018} use machine learning to decide if linearizing
		the problem will solve faster.
		When the \gls{qp} problem given by the relaxation is convex, \textit{i.e.}, the quadratic objective
		matrix is semidefinite positive, one could address the problem by a \gls{bnb} algorithm that solves
		\gls{qp} relaxations\footnote{
		    Note that convex \glspl{qp} can be solved in polynomial time.
		} to provide lower bounds.
		Even in this convex case, it is not clear if \gls{qp} \gls{bnb} would solve faster than linearizing
		the problem (by using \cite{McCormickComputabilityglobalsolutions1976} inequalities) and solving an
		equivalent \gls{milp}.
		This is why \gls{ml} is a great candidate here to fill the knowledge gap.
		In both papers \citep{KruberLearningWhenUse2017, BonamiLearningClassificationMixedInteger2018}, the
		authors experiment with different \gls{ml} models, such as support vector machines and random
		forests, as is good practice when no prior knowledge is embedded in the model.

		The heuristic building framework used in \cite{KarapetyanMarkov2017} and \cite{MasciaGrammar2014},
		already presented in Section~\ref{sec:experience}, can be understood under this eye.
		Indeed, it can be seen as a large parametric heuristic, configured by the transition probabilities
		in the former case, and by the parameter representing a sentence in the latter.

		As previously stated, the parametrization of the \gls{co} algorithm provided by \gls{ml} is to be
		understood in a very broad sense.
		For instance, in the case of radiation therapy for cancer treatment,
		\cite{MahmoodAutomatedTreatmentPlanning2018} use \gls{ml} to produce candidate therapies that are
		afterward refined by a \gls{co} algorithm into a deliverable plan.
		Namely, a \gls{gan} is used to color CT scan images into a potential radiation plan, then, inverse
		optimization \citep{AhujaInverseOptimization2001} is applied on the result to make the plan
		feasible \citep{ChanGeneralizedInverseMultiobjective2014}.
		In general, \glspl{gan} are made of two distinct networks: one (the generator) generates images,
		and another one (the discriminator) discriminates between the generated images and a dataset of
		real images.
		Both are trained alternatively: the discriminator through a usual supervised objective, while the
		generator is trained to fool the discriminator.
		In \cite{MahmoodAutomatedTreatmentPlanning2018}, a particular type of \gls{gan} (conditional
		\gls{gan}) is used to provide coloring instead of random images.
		The interested reader is referred to \cite{CreswellGenerativeAdversarialNetworks2018} for an
		overview on \glspl{gan}.

		We end this section by noting that an \gls{ml} model used for learning some representation may in
		turn use as features pieces of information given by another \gls{co} algorithm, such as the
		decomposition statistics used in \cite{KruberLearningWhenUse2017}, or the \gls{lp} information in
		\cite{BonamiLearningClassificationMixedInteger2018}.
		Moreover, we remark that, in the satisfiability context, the learning of the type of algorithm
		to execute on a particular cluster of instances has been paired with the learning of the
		parameters of the algorithm itself, see, \textit{e.g.},
		\cite{AnsoteguiReactive2017, AnsoteguiHyperReactive2019}.

		\subsubsection{Machine learning alongside optimization algorithms}
		\label{sec:learning_alongside}
\begin{figure}[H]
			\centering
			\includegraphics[scale=.85]{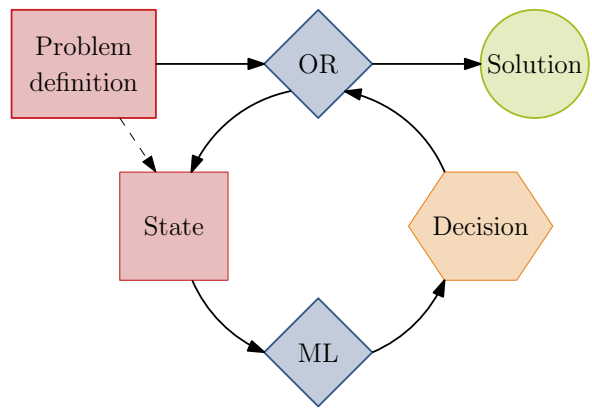}
			\caption{The \acrlong{co} algorithm repeatedly queries the same \gls{ml} model to
				make decisions.
				The \gls{ml} model takes as input the current state of the algorithm, which may include the
				problem definition.
			}
			\label{fig:ml_or_alongside}
		\end{figure}

		To generalize the context of the previous section to its full potential, one can build \gls{co}
		algorithms that repeatedly call an \gls{ml} model throughout their execution, as illustrated in
		Figure~\ref{fig:ml_or_alongside}.
		A master algorithm controls the high-level structure while frequently calling an \gls{ml} model to
		assist in lower level decisions.
		The key difference between this approach and the examples discussed in the previous section is that
		the \textit{same \gls{ml} model} is used by the \gls{co} algorithm to make the same type of
		decisions a number of times in the order of the number of iterations of the algorithm.
		As in the previous section, nothing prevents one from applying additional steps before or after
		such an algorithm.

		This is clearly the context of the \acrlong{bnb} tree for \gls{milp}, where we already mentioned
		how the task of selecting the branching variable is either too heuristic or too slow, and is
		therefore a good candidate for learning \citep{Lodilearningbranchingsurvey2017}.
		In this case, the general algorithm remains a \acrlong{bnb} framework, with the same software
		architecture and the same guarantees on lower and upper bounds, but the branching decisions made at
		every node are left to be learned.
		Likewise, the work of \cite{HottungDeep2017} learning both a branching policy \emph{and} value
		network for heuristic tree search undeniably fits in this context.
		Another important aspect in solving \glspl{milp} is the use of primal heuristics, \textit{i.e.},
		algorithms that are applied in the \gls{bnb} nodes to find feasible solutions, without guarantee of
		success.
		On top of their obvious advantages, good solutions also give tighter upper bounds (for minimization
		problems) on the solution value and make more pruning of the tree possible.
		Heuristics depend on the branching node (as branching fix some variables to specific values), so
		they need to be run frequently.
		However, running them too often can slow down the exploration of the tree, especially if their
		outcome is negative, \textit{i.e.}, no better upper bound is detected.
		\cite{KhalilLearningRunHeuristics2017} build an \gls{ml} model to predict whether
		or not running a given heuristic will yield a better solution than the best one found so far and
		then greedily run that heuristic whenever the outcome of the model is positive.

		The approximation used by \cite{Baltean-LugojanStrongsparsecut2018}, already discussed in Section
		\ref{sec:end2end_learning}, is an example of predicting a high-level description of the solution to
		an optimization problem, namely the objective value.
		Nonetheless, the goal is to solve the original \gls{qp}.
		Thus, the learned model is queried repeatedly to select promising cutting planes.
		The \gls{ml} model is used only to select promising cuts, but once selected, cuts are added to the
		\gls{lp} relaxation, thus embedding the \gls{ml} outcome into an exact algorithm.
		This approach highlights promising directions for this type of algorithm.
		The decision learned is critical because adding the best cutting planes is necessary for solving
		the problem fast (or fast enough, because in the presence of NP-hard problems, optimization may
		time out before any meaningful solving).
		At the same time, the approximate decision (often in the form of a probability) does not compromise
		the exactness of the algorithm: any cut added is guaranteed to be valid.
		This setting leaves room for \gls{ml} to thrive, while reducing the need for guarantees from the
		\gls{ml} algorithms (an active and difficult area of research).
		In addition, note that, the approach in \cite{LarsenPredictingSolutionSummaries2018} is part of a
		master algorithm in which the \gls{ml} is iteratively invoked to make booking decisions in real
		time.
		The work of \cite{KhalilLearningCombinatorialOptimization2017}, presented in Section
		\ref{sec:experience}, also belongs to this setting, even if the resulting algorithm is heuristic.
		Indeed, an \gls{ml} model is asked to select the most relevant node, while a master
		algorithm maintains the partial tour, computes its length, \textit{etc}.
		Because the master algorithm is very simple, it is possible to see the contribution as an
		end-to-end method, as stated in Section~\ref{sec:end2end_learning}, but it can also be interpreted
		more generally as done here.

		Presented in Section~\ref{sec:experience}, and mentioned in the previous section, the Markov Chain
		framework for building heuristics from \cite{KarapetyanMarkov2017} can also be framed as repeated
		decisions.
		The transition matrix can be queried and sampled from in order to transition from one state to
		another, \textit{i.e.}, to make the low level decisions of choosing the next move.
		The three distinctions made in this Section~\ref{sec:algorithmic_structure} are general enough that
		they can overlap.
		Here, the fact that the model operates on internal state transitions, yet is learned globally, is
		what makes it hard to analyze.

		Before ending this section, it is worth mentioning that learning recurrent algorithmic decisions is
		also used in the deep learning community, for instance in the field of meta-learning to decide how
		to apply gradient updates in stochastic gradient descent
		\citep{AndrychowiczLearninglearngradient2016, LiLearningOptimizeNeural2017,
		WichrowskaLearnedOptimizersthat2017}.

\section{Learning objective}
	\label{sec:learning_objective}

	In the previous section, we have surveyed the existing literature by orthogonally grouping the main
	contributions of \gls{ml} for \gls{co} into families of approaches, sometimes with overlaps.
	In this section, we formulate and study the objective that drives the learning process.

	\subsection{Multi-instance formulation}
		\label{sec:formulation}
		In the following, we introduce an abstract learning formulation (inspired from
		\cite{BischlAslib2016}).
		How would an \gls{ml} practitioner compare optimization algorithms?
		Let us define $\mathcal{I}$ to be a set of problem instances, and $P$ a probability distribution
		over $\mathcal{I}$.
		These are the problems that we care about, weighted by a probability distribution, reflecting the
		fact that, in a practical application, not all problems are as likely.
		In practice, $\mathcal{I}$ or $P$ are inaccessible, but we can observe some samples from $P$, as motivated in
		the introduction with the Montreal delivery company.
		For a set of algorithms $\mathcal{A}$, let $m: \mathcal{I} \times \mathcal{A} \rightarrow
		\mathbb{R}$ be a measure of the performance of an algorithm on a problem instance (lower is better).
		This could be the objective value of the best solution found, but could also incorporate
		elements from optimality bounds, absence of results, running times, and resource usage.
		To compare $a_{1},\ a_2 \in \mathcal{A}$, an \gls{ml} practitioner would compare $\mathbb{E}_{i \sim
				P}\ m(i, a_1)$ and $\mathbb{E}_{i \sim P}\ m(i, a_2)$, or equivalently
		\begin{equation}
			\label{eq:model_comparison}
			\min_{a \in \{a_1, a_2\}} \mathbb{E}_{i \sim P}\ m(i, a).
		\end{equation}
		Because measuring these quantities is not tractable, one will typically use empirical
		estimates instead, by using a finite dataset $D_{train}$ of independent instances
		sampled from $P$
		\begin{equation}
		    \label{eq:model_comparison_empirical}
			\min_{a \in \{a_1, a_2\}} \sum_{i \in D_{train}} \frac{1}{|D_{train}|} m(i, a).
		\end{equation}
		This is intuitive and done in practice: collect a dataset of problem instances and compare say,
		average running times.
		Of course, such expectation can be computed for different datasets (different
		$\mathcal{I}$'s and $P$'s), and different measures (different $m$'s).

		This is already a learning problem.
		The more general one that we want to solve through leaning is
		\begin{equation}
			\label{eq:learning_pb}
			\min_{a \in \mathcal{A}} \mathbb{E}_{i \sim P}\ m(i, a).
		\end{equation}
		Instead of comparing between two algorithms, we may compare among an uncountable, maybe
		non-parametric, space of algorithms.
		To see how we come up with so many algorithms, we have to look at the algorithms in
		Section~\ref{sec:recent_approaches}, and think of the \gls{ml} model space over which we learn as
		defining parametrizing the algorithm space $\mathcal{A}$.
		For instance, consider the case of learning a branching policy $\pi$ for \gls{bnb}.
		If we define the policy to be a neural network with a set of weights $\theta \in \mathbb{R}^p$,
		then we obtain a parametric \gls{bnb} algorithm $a(\pi_\theta)$ and \eqref{eq:learning_pb} becomes
		\begin{equation}
			\min_{\theta \in \mathbb{R}^p} \mathbb{E}_{i \sim P}\ m(i, a(\pi_\theta)).
		\end{equation}

		Unfortunately, solving this problem is hard.
		On the one hand, the performance measure $m$ is most often not differentiable and without
		closed form expression.
		We discuss this in Section~\ref{sec:learning_proxies}.
		On the other hand, computing the expectation in \eqref{eq:learning_pb} is intractable.
		As in \eqref{eq:model_comparison_empirical}, one can use an empirical distribution using a
		finite dataset, but that leads to \emph{generalization} considerations, as explained in
		Section~\ref{sec:generalization}.

		Before we move on, let us introduce a new element to make \eqref{eq:learning_pb} more
		general.
		That formula suggests that, once given an instance, the outcome of the performance measure is
		deterministic.
		That is unrealistic for multiple reasons.
		The performance measure could itself incorporate some source of randomness due to external
		factors, for instance with running times which are hardware and system dependent.
		The algorithm could also incorporate non negligible sources of randomness, if it is designed
		to be stochastic, or if some operations are non deterministic, or to express the fact that the
		algorithm should be robust to the choice of some external parameters.
		Let $\tau$ be that source of randomness, $\pi \in \Pi$ the internal policy being learned, and
		$a(\pi, \tau)$ the resulting algorithm, then we can reformulate \eqref{eq:learning_pb} as
		\begin{equation}
			\label{eq:learning_pb_random}
			\min_{\pi \in \Pi}\
			\mathbb{E}_{i \sim P}
			\left [ \,
			    \mathbb{E}_\tau [\, m(i,\, a(\pi, \tau))\ |\ i \,]
			\, \right ].
		\end{equation}
		In particular, when learning repeated decisions, as in
		Section~\ref{sec:learning_alongside}, this source of randomness can be expressed along the
		trajectory followed in the \gls{mdp}, using the dynamics of the environment $p(s',r|a,s)$ (see
		Figure~\ref{fig:mdp}).
		The addition made in \eqref{eq:learning_pb_random} will be useful for the discussion on
		generalization in Section~\ref{sec:generalization}.

	\subsection{Surrogate objectives}
		\label{sec:learning_proxies}
		In the previous section, we have formulated a proper learning objective.
		Here, we try to relate that objective to the learning methods of
		Section~\ref{sec:learning_methods}, namely, demonstration and experience.
		If the usual learning metrics of an \gls{ml} model, \textit{e.g.}, accuracy for classification in
		supervised (imitation) learning, is improving, does it mean that the performance metric of
		\eqref{eq:learning_pb} is also improving?

		A straightforward approach for solving \eqref{eq:learning_pb_random} is that of reinforcement
		learning (including direct optimization methods), as surveyed in Section~\ref{sec:experience}.
		The objective from \eqref{eq:learning_pb} can be optimized directly on experience data by
		matching the total return to the performance measure.
		Sometimes, a single final reward can naturally be decoupled across the trajectory.
		For instance, if the performance objective of a \gls{bnb} variable selection policy is to minimize
		the number of opened nodes, then the policy can receive a reward discouraging an increase in the
		number of nodes, hence  giving an incentive to select variables that lead to pruning.
		However, that may not be always possible, leaving only the option of delaying a single reward to
	    the end of the trajectory.
		This sparse reward setting is challenging for \gls{rl} algorithms, and one might want to design a
		surrogate reward signal to encourage intermediate accomplishments.
		This introduces some discrepancies, and the policy being optimized may learn a behavior not
		intended by the algorithm designer.
		There is \textit{a priori} no relationship between two reward signals.
		One needs to make use of their intuition to design surrogate signals, \textit{e.g.}, minimizing the
		number of \gls{bnb} nodes \textit{should} lead to smaller running times.
		Reward shaping is an active area of research in \gls{rl}, yet it is often performed by a number of
		engineering tricks.

		In the case of learning a policy from a supervised signal from expert demonstration, the
		performance measure $m$ does not even appear in the learning problem that is solved.
		In this context, the goal is to optimize a policy $\pi \in \Pi$ in the action space to mimic an
		expert policy $\pi_e$ (as first introduced with Figure~\ref{fig:demonstration})
		\begin{equation}
			\label{eq:behavior_cloning}
			\min_{\pi \in \Pi}\ \mathbb{E}_{i \sim P}
			\left [ \,
			    \mathbb{E}_{s} [\, \ell(\pi(s), \pi_e(s))\ |\ i, \pi_e \,]
			\, \right ],
		\end{equation}
		where $\ell$ is a task dependent loss (classification, regression, \textit{etc.}).
		We have emphasized that the state $S$ is conditional, not only on the instance, but also on the
		expert policy $\pi_e$ used to collect the data.
		Intuitively, the better the \gls{ml} model learns, \textit{i.e.}, the better the policy imitates
		the expert, the closer the final performance of the learned policy should be to the performance of
		the expert.
		Under some conditions, it is possible to relate the performance of the learned policy to the
		performance of the expert policy, but covering this aspect is out of the scope of this paper.
		The opposite is not true, if learning fails, the policy may still turn out to perform well (by
		encountering an alternative good decision).
		Indeed, when making a decision with high surrogate objective error, the learning will be fully
		penalized when, in fact, the decision could have good performances by the original metric.
		For that reason, it is capital to report the performance metrics.
		For example, we surveyed in Section~\ref{sec:algorithm_configuration} the work of
		\cite{BonamiLearningClassificationMixedInteger2018} where the authors train a classifier to predict
		if a mixed integer quadratic problem instance should be linearized or not.
		The targets used for the learner are computed optimally by solving the problem instance in both
		configurations.
		Simply reporting the classification accuracy is not enough.
		Indeed, this metric gives no information on the impact a misclassification has on running times,
		the metric used to compute the targets.
		In the binary classification, a properly classified example could also happen to have unsignificant
		difference between the running times of the two configurations.
		To alleviate this issue, the authors also introduce a category where running times are not
		significatively different (and report the real running times).
		A continuous extension would be to learn a regression of the solving time.
		However, learning this regression now means that the final algorithm needs to optimize over the set
		of decisions to find the best one.
		In \gls{rl}, this is analoguous to learning a value function (see Section~\ref{sec:maching_learning}).
		Applying the same reasoning to repeated decisions is better understood with the complete \gls{rl}
		theory.

	\subsection{On generalization}
		\label{sec:generalization}
		In Section~\ref{sec:formulation}, we have claimed that the probability distribution in
		\eqref{eq:learning_pb} is inaccessible and needs to be replaced by the empirical probability
		distribution over a finite dataset $D_{train}$.
		The optimization problem solved is
		\begin{equation}
			\label{eq:learning_pb_empirical}
			    \min_{a \in \mathcal{A}}\ \sum_{i \in D_{train}} \frac{1}{|D_{train}|} m(i, a).
		\end{equation}
		As pointed out in Section~\ref{sec:maching_learning}, when optimizing over
		the empirical probability distribution, we risk having a low performance measure on the finite
		number of problem instances, \emph{regardless of the true probability distribution}.
		In this case, the \emph{generalization} error is high because of the discrepancy between the
		training performances and the true expected performances (overfitting).
		To control this aspect, a validation set $D_{valid}$ is introduced to compare a finite number of
		candidate algorithms based on estimates of generalization performances, and a test set $D_{test}$
		is used for estimating the generalization performances of the selected algorithm.

		In the following, we look more intuitively at generalization in \gls{ml} for \gls{co}, and its
		consequences.
		To make it easier, let us recall different learning scenarios.
		In the introduction, we have motivated the Montreal delivery company example, where the problems of
		interest are from an unknown probability distribution of Montreal \glspl{tsp}.
		This is a very restricted set of problems, but enough to deliver value for this business.
		Much more ambitious, we may want our policy learned on a finite set of instances to perform well
		(generalize) to any ``\textit{real-world}'' \gls{milp} instance.
		This is of interest if you are in the business of selling \gls{milp} solvers and want the branching
		policy to perform well for as many of your clients as possible.
		In both cases, generalization applies to the instances that are not known to the algorithm
		implementer.
		These are the only instances that we care about; the one used for training are already solved.
		The topic of probability distribution of instances also appears naturally in stochastic
		programming/optimization, where uncertainty about the problem is modeled through probability
		distributions.
		Scenario generation, an essential way to solve this type of optimization programs, require sampling
		from this distribution and solving the associated problem multiple times.
		\cite{NairLearning2018} take advantage of this repetitive process to learn an end-to-end
		model to solve the problem.
		Their model is composed of a local search and a local improvement policy and is trained through
		\gls{rl}.
		Here, generalization means that, during scenario generation, the learned search beats other
		approaches, hence delivering an overall faster stochastic programming algorithm.
		In short, \emph{learning without generalization is pointless!}

		When the policy generalizes to other problem instances, it is no longer a problem if training
		requires additional computation for solving problem instances because, learning can be decoupled
		from solving as it can be done offline.
		This setting is promising as it could give a policy to use out of the box for similar instances,
		while keeping the learning problem to be handled beforehand while remaining hopefully reasonable.
		When the model learned is a simple mapping, as is the case in
		Section~\ref{sec:algorithm_configuration}, generalization to new instances, as previously
		explained, can be easily understood.
		However, when learning sequential decisions, as in Section~\ref{sec:learning_alongside}, there are
		intricate levels of generalization.
		We said that we want the policy to generalize to new instances, but the policy also needs to
		generalize to internal states of the algorithm for a single instance, even if the model can be
		learned from complete optimization trajectories, as formulated by
		\eqref{eq:learning_pb_random}.
		Indeed, complex algorithms can have unexpected sources of randomness, even if they are designed to
		be deterministic.
		For instance, a numerical approximation may perform differently if the version of some underlying
		numerical library is changed or because of asynchronous computing, such as when using Graphical
		Processing Units \citep{NagarajanDeterministic2019}.
		Furthermore, even if we can achieve perfect replicability, we do not want the branching
		policy to break if some other parameters of the solver are set (slightly) differently.
		At the very least, we want the policy to be robust to the choice of the random seed present in many
		algorithms, including \gls{milp} solvers.
		These parameters can therefore be modeled as random variables.
		Because of these nested levels of generalization, one appealing way to think about the training data
		from multiple instances is like separate tasks of a multi-task learning setting.
		The different tasks have underlying aspects in common, and they may also have their own peculiar
		quirks.
		One way to learn a single policy that generalizes within a distribution of instances is to take
		advantage of these commonalities.
		Generalization in \gls{rl} remains a challenging topic, probably because of the fuzzy distinction
		between a multi-task setting, and a large environment encompassing all of the tasks.

		Choosing how ambitious one should be in defining the characteristics of the distribution is a hard
		question.
		For instance, if the Montreal company expands its business to other cities, should they be
		considered as separate distributions, and learn one branching policy per city, or only a single one?
		Maybe one per continent?
		Generalization to a larger variety of instances is challenging and requires more advanced and
		expensive learning algorithms.
		Learning an array of \gls{ml} models for different distributions asociated with a same task
		means of course more models to train, maintain, and deploy.
		The same goes with traditional \gls{co} algorithms, an \gls{milp} solver on its own is not the best
		performing algorithm to solve \glspl{tsp}, but it works across all \gls{milp} problems.
		It is too early to provide insights about how broad the considered distributions should be,
		given the limited literature in the field.
		For scholars generating synthetic distributions, two intuitive axes of investigation are
		``\textit{structure}'' and ``\textit{size}''.
		A \gls{tsp} and a scheduling problem seem to have fairly different structure, and one can think of
		two planar euclidean \glspl{tsp} to be way more similar.
		Still, two of these \glspl{tsp} can have dramatically different sizes (number of nodes).
		For instance, \cite{GasseExact2019} assess their methodology independently on three distributions.
		Each training dataset has a specific problem structure (set covering, combinatorial auction, and
		capacitated facility location), and a fixed problem size.
		The problem instance generators used are state-of-art and representative of real-world instances.
		Nonetheless, when they evaluate their learned algorithm, the authors push the test distributions to
		larger sizes.
		The idea behind this is to gauge if the model learned is able to generalize to a larger, more
		practical, distribution, or only perform well on the restricted distribution of problems of the same
		size.
		The answer is largely affirmative.

	\subsection{Single instance learning}
		An edge case that we have not much discussed yet is the single instance learning framework.
		This might be the case for instance for planning the design of a single factory.
		The factory would only be built once, with very peculiar requirements, and the planners are not
		interested to relate this to other problems.
		In this case, one can make as many runs (episodes) and as many calls to a potential expert
		or simulator as one wants, but ultimately one only cares about solving this one instance.
		Learning a policy for a single instance should require a simpler \gls{ml} model, which could thus
		require less training examples.
		Nonetheless, in the single instance case, one learns the policy from scratch at every new instance,
		actually incorporating learning (not learned models but really the learning process itself) into
		the end algorithm.
		This means starting the timer at the beginning of learning and competing with other solvers to get
		the solution the fastest (or get the best results within a time limit).
		This is an edge scenario that can only be employed in the setting of the
		Section~\ref{sec:learning_alongside}, where \gls{ml} is embedded \emph{inside} a \gls{co}
		algorithm; otherwise there would be only one training example!
		There is therefore no notion of generalization to other problem instances, so
		\eqref{eq:learning_pb} is not the learning problem being solved.
		Nonetheless, the model still needs to generalize to \emph{unseen states} of the algorithm.
		Indeed, if the model was learned from all states of the algorithm that are needed to solve the
		problem, then the problem is already solved at training time and learning is therefore fruitless.
		This is the methodology followed by \cite{KhalilLearningbranchmixed2016}, introduced in
		Section~\ref{sec:demonstration}, to learn an instance-specific branching policy.
		The policy is learned from strong-branching at the top of the \gls{bnb} tree, but needs to
		generalize to the state of the algorithm at the bottom of the tree, where it is used.
		However, as for all \gls{co} algorithms, a fair comparison to another algorithm can only be done on
		an independent dataset of instances, as in \eqref{eq:model_comparison}.
		This is because through human trials and errors, the data used when building the algorithm leaks
		into the design of the algorithm, even without explicit learning components.

	\subsection{Fine tuning and meta-learning}
		A compromise between instance-specific learning and learning a generic policy is what we typically
		have in multi-task learning: some parameters are shared across tasks and some are specific to each
		task.
		A common way to do that (in the transfer learning scenario) is to start from a generic policy and
		then adapt it to the particular instance by a form of \emph{fine-tuning} procedure: training
		proceeds in two stages, first training the generic policy across many instances from the same
		distribution, and then continuing training on the examples associated with a given instance on
		which we are hoping to get more specialized and accurate predictions.

		Machine learning advances in the areas of meta-learning and transfer learning are particularly
		interesting to consider here.
		Meta-learning considers two levels of optimization: the inner loop trains the parameters of a model
		on the training set in a way that depends on meta-parameters, which are themselves optimized in an
		outer loop (\textit{i.e.}, obtaining a gradient for each completed inner-loop training or update).
		When the outer loop's objective function is performance on a validation set, we end up training a
		system so that it will generalize well.
		This can be a successful strategy for generalizing from very few examples if we have access to many
		such training tasks.
		It is related to transfer learning, where we want that what has been learned in one or many tasks
		helps improve generalization on another.
		These approaches can help rapidly adapt to a new problem, which would be useful in the context of
		solving many \gls{milp} instances, seen as many related tasks.

		To stay with the branching example on \glspl{milp}, one may not want the policy to perform well out
		of the box on new instances (from the given distribution).
		Instead, one may want to learn a policy offline that can be adapted to a new instance in a few
		training steps, every time it is given one.
		Similar topics have been explored in the context of automatic configuration tools.
		\cite{FitzgeraldReact2014} study the automatic configuration in the lifelong learning
		context (a form of sequential transfer learning).
		The automatic configuration algorithm is augmented with a set of previous configurations that are
		prioritized on any new problem instance.
		A score reflecting past performances is kept along every configuration.
		It is designed to retain configurations that performed well in the past, while letting
		new ones a chance to be properly evaluated.
		The automatic configuration optimization algorithm used by \cite{LindauerWarmstarting2018} requires
		training an empirical cost model mapping the Cartesian product of parameter configurations and
		problem instances to expected algorithmic performance.
		Such a model is usually learned for every cluster of problem instance that requires configuring.
		Instead, when presented with a new cluster, the authors combine the previously learned cost models
		and the new one to build an ensemble model.
		As done by \cite{FitzgeraldReact2014}, the authors also build a set of previous configurations to
		prioritize, using an empirical cost model to fill the missing data.
		This setting, which is more general than not performing any adaptation of the policy, has potential
		for better generalization.
		Once again, the scale on which this is applied can vary depending on ambition.
		One can transfer on very similar instances, or learn a policy that transfers to a vast range of
		instances.

		Meta-learning algorithms were first introduced in the 1990s \citep{BengioLearningSynaptic1991,
		SchmidhuberLearningControl1992, ThrunPrattBook1998} and have since then become particularly popular
		in \gls{ml}, including, but not limited to, learning a gradient update rule
		\citep{HochreiterLearningLearn2001, AndrychowiczLearninglearngradient2016}, few shot learning
		\citep{RaviOptimizationFewShot2017}, and multi-task \gls{rl}
		\citep{FinnModelAgnosticMetaLearningFast2017}.

	\subsection{Other metrics}
		Other metrics from the process of learning itself are also relevant, such as how fast the learning
		process is, the sample complexity (number of examples required to properly fit the model),
		\textit{etc}.
		As opposed to the metrics suggested earlier in this section, these metrics provide us with
		information not about final performance, but about offline computation or the number of training
		examples required to obtain the desired policy.
		This information is, of course, useful to calibrate the effort in integrating \gls{ml} into
		\gls{co} algorithms.

\section{Methodology}
	\label{sec:methodology}
	In the previous section, we have detailed the theoretical learning framework of using \gls{ml} in
	\gls{co} algorithms.
	Here, we provide some additional discussion broadening some previously made claims.

	\subsection{Demonstration and experience}
		\label{sec:demonstration_and_experience}
		In order to learn a policy, we have highlighted two methodologies: demonstration, where the
		expected behavior is shown by an expert or oracle (sometimes at a considerable computational cost),
		and experience, where the policy is learned through trial and error with a reward signal.

		In the demonstration setting, the performance of the learned policy is bounded by the expert, which
		is a limitation when the expert is not optimal.
		More precisely, without a reward signal, the imitation policy can only hope to marginally
		outperform the expert (for example because the learner can reduce the variance of the answers across similarly-performing experts).
		The better the learning, the closer the performance of the learner to the expert's.
		This means that imitation alone should be used only if it is significantly faster than the expert
		to compute the policy.
		Furthermore, the performance of the learned policy may not generalize well to unseen examples and
		small variations of the task and may be unstable due to accumulation of errors.
		This is because in \eqref{eq:behavior_cloning}, the data was collected according to the
		expert policy $\pi_e$, but when run over multiple repeated decisions, the distribution of states
		becomes that of the learned policy.
		Some downsides of supervised (imitation) learning can be overcome with more advanced algorithms,
		including active methods to query the expert as an oracle to improve behavior in uncertain states.
		The part of imitation learning presented here is limited compared to the current literature
		in \gls{ml}.

		On the contrary, with a reward, the algorithm learns to optimize for that signal and can
		potentially outperform any expert, at the cost of a much longer training time.
		Learning from a reward signal (experience) is also more flexible when multiple decisions are
		(almost) equally good in comparison with an expert that would favor one (arbitrary) decision.
		Experience is not without flaws.
		In the case where policies are approximated (\textit{e.g.}, with a neural network), the learning
		process may get stuck around poor solutions if exploration is not sufficient or solutions which do
		not generalize well are found.
		Furthermore, it may not always be straightforward to define a reward signal.
		For instance, sparse rewards may be augmented using reward shaping or a curriculum in order to
		value intermediate accomplishments (see Section \ref{sec:maching_learning}).

		Often, it is a good idea to start learning from demonstrations by an expert, then refine the policy
		using experience and a reward signal.
		This is what was done in the original AlphaGo paper \citep{SilverMasteringgameGo2016}, where human
		knowledge is combined with reinforcement learning.
		The reader is referred to \cite{HusseinImitationLearningSurvey2017} for a survey on imitation
		learning covering most of the discussion in this section.

	\subsection{Partial observability}
		\label{sec:partial_observability}
		We mentioned in section \ref{sec:maching_learning} that sometimes the states of an \gls{mdp} are
		not fully observed and the Markov property does not hold, \textit{i.e.}, the probability of the
		next observation, conditioned on the current observation and action, is not equal to the
		probability of the next observation, conditioned on all past observations and actions.
		An immediate example of this can be found in any environment simulating physics: a single
		frame/image of such an environment is not sufficient to grasp notions such as velocity and is
		therefore not sufficient to properly estimate the future trajectory of objects.
		It turns out that, on real applications, partial observability is closer to the norm than to the
		exception, either because one does not have access to a true state of the environment, or because
		it is not computationally tractable to represent and needs to be approximated.
		A straightforward way to tackle the problem is to compress all previous observations using an
		\gls{rnn}.
		This can be applied in the imitation learning setting, as well as in \gls{rl}, for instance by
		learning a recurrent policy \citep{WierstraRecurrentpolicy2010}.

		How does this apply in the case where we want to learn a policy function making decisions for a
		\gls{co} algorithm?
		On the one hand, one has full access to the state of the algorithm because it is represented in
		exact mathematical concepts, such as constraints, cuts, solutions, \gls{bnb} tree, \textit{etc}.
		On the other hand, these states can be exponentially large.
		This is an issue in terms of computations and generalization.
		Indeed, if one does want to solve problems quickly, one needs to have a policy that is also fast to
		compute, especially if it is called frequently as is the case for, say, branching decisions.
		Furthermore, considering too high-dimensional states is also a statistical problem for learning, as
		it may dramatically increase the required number of samples, decrease the learning speed, or
		fail altogether.
		Hence, it is necessary to keep these aspects in mind while experimenting with different
		representations of the data.

	\subsection{Exactness and approximation}
		In the different examples we have surveyed, \gls{ml} is used in both exact and heuristic
		frameworks, for example \cite{Baltean-LugojanStrongsparsecut2018} and
		\cite{LarsenPredictingSolutionSummaries2018}, respectively.
		Getting the output of an \gls{ml} model to respect advanced types of constraints is a hard task.
		In order to build exact algorithms with \gls{ml} components, it is necessary to apply the \gls{ml}
		where all possible decisions are valid.
		Using only \gls{ml} as surveyed in Section~\ref{sec:end2end_learning} cannot give any optimality
		guarantee, and only weak feasibility guarantees (see Section~\ref{sec:feasibility}).
		However, applying \gls{ml} to select or parametrize a \gls{co} algorithm as in Section
		\ref{sec:algorithm_configuration} will keep exactness if all possible choices that \gls{ml}
		discriminate lead to complete algorithms.
		Finally, in the case of repeated interactions between \gls{ml} and \gls{co} surveyed in Section
		\ref{sec:learning_alongside}, all possible decisions must be valid.
		For instance, in the case of \glspl{milp}, this includes branching \emph{among fractional
		variables} of the \gls{lp} relaxation, selecting the node to explore \emph{among open branching
		nodes} \citep{HeLearningSearchBranch2014}, deciding on the frequency to run heuristics on the
		\gls{bnb} nodes \citep{KhalilLearningRunHeuristics2017}, selecting cutting planes \emph{among valid
		inequalities} \citep{Baltean-LugojanStrongsparsecut2018}, removing previous cutting planes
		\emph{if they are not original constraints or branching decision}, \textit{etc}.
		A counter-example can be found in the work of \cite{HottungDeep2017}, presented in
		Section~\ref{sec:demonstration}.
		In their branch-an-bound framework, bounding is performed by an approximate \gls{ml} model that can
		overestimate lower bounds, resulting in invalid pruning.
		The resulting algorithm is therefore not an exact one.

\section{Challenges}
	\label{sec:challenges}
	In this section, we are reviewing some of the algorithmic concepts previously introduced by taking
	the viewpoint of their associated challenges.

	\subsection{Feasibility}
		\label{sec:feasibility}
		In Section~\ref{sec:end2end_learning}, we pointed out how \gls{ml} can be used to directly output
		solutions to optimization problems.
		Rather than learning the solution, it would be more precise to say that the algorithm is learning a
		\textit{heuristic}.
		As already repeatedly noted, the learned algorithm does not give any guarantee in terms of
		optimality, but it is even more critical that feasibility is not guaranteed either.
		Indeed, we do not know how far the output of the heuristic is from the optimal solution, or if it
		even respects the constraints of the problem.
		This can be the case for every heuristic and the issue can be mitigated by using the heuristic
		within an exact optimization algorithm (such as branch and bound).

		Finding feasible solutions is not an easy problem (theoretically NP-hard for \glspl{milp}), but it
		is even more challenging in \gls{ml}, especially by using neural networks.
		Indeed, trained with gradient descent, neural architectures must be designed carefully in order not
		to break differentiability.
		For instance, both pointer networks \citep{VinyalsPointerNetworks2015} and the Sinkhorn layer
		\citep{EmamiLearningPermutationsSinkhorn2018} are complex architectures used to make a network
		output a permutation, a constraint easy to satisfy when writing a classical \gls{co} heuristic.

	\subsection{Modelling}
		\label{sec:modelling}
		In \gls{ml}, in general, and in deep learning, in particular, we know some good prior for some
		given problems.
		For instance, we know that a \gls{cnn} is an architecture that will learn and generalize more
		easily than others on image data.
		The problems studied in \gls{co} are different from the ones currently being addressed in \gls{ml},
		where most successful applications target natural signals.
		The architectures used to learn good policies in \acrlong{co} might be very different from what is
		currently used with deep learning.
		This might also be true in more subtle or unexpected ways: it is conceivable that, in turn, the
		optimization components of deep learning algorithms (say, modifications to \gls{sgd}) could be
		different when deep learning is applied to the \gls{co} context.

		Current deep learning already provides many techniques and architectures for tackling problems of
		interest in \gls{co}.
		As pointed out in section \ref{sec:maching_learning}, techniques such as parameter sharing made it
		possible for neural networks to process sequences of variable length with \glspl{rnn} or, more
		recently, to process graph structured data through \glspl{gnn}.
		Processing graph data is of uttermost importance in \gls{co} because many problems are formulated
		(represented) on graphs.
		For a very general example, \cite{SelsamLearningSATSolver2018} represent a satisfiability problem
		using a bipartite graph on variables and clauses.
		This can generalize to \glspl{milp}, where the constraint matrix can be represented as the
		adjacency matrix of a bipartite graph on variables and constraints, as done in
		\cite{GasseExact2019}.

	\subsection{Scaling}
		Scaling to larger problems can be a challenge.
		If a model trained on instances up to some size, say \glspl{tsp} up to size fifty nodes, is
		evaluated on larger instances, say \glspl{tsp} of size a hundred, five hundred nodes, \textit{etc},
		the challenge exists in terms of generalization, as mentioned in Section~\ref{sec:generalization}.
		Indeed, all of the papers tackling \gls{tsp} through \gls{ml} and attempting to solve larger
		instances see degrading performance as size increases much beyond the sizes seen during training \citep{VinyalsPointerNetworks2015,
	    BelloNeuralCombinatorialOptimization2016, KhalilLearningCombinatorialOptimization2017,
		KoolAttentionSolvesYour2018}.
		To tackle this issue, one may try to learn on larger instances, but this may turn out to be a
		computational and generalization issue.
		Except for very simple \gls{ml} models and strong assumptions about the data distribution, it is
		impossible to know the computational complexity and the sample complexity, \text{i.e.} the number
		of observations that learning requires, because one is unaware of the exact problem one is trying
		to solve, \textit{i.e.}, the true data generating distribution.

	\subsection{Data generation}
		Collecting data (for example instances of optimization problems) is a subtle task.
		\cite{LarsenPredictingSolutionSummaries2018} claim that ``\textit{sampling from historical
		data is appropriate when attempting to mimic a behavior reflected in such data}''.
		In other words, given an external process on which we observe instances of an optimization problem,
		we can collect data to train some policy needed for optimization, and expect the policy to
		generalize on future instances of this application.
		A practical example would be a business that frequently encounters optimization problems related to
		their activities, such as the Montreal delivery company example used in the introduction.

		In other cases, \textit{i.e.}, when we are not targeting a specific application for which we would
		have historical data, how can we proactively train a policy for problems that we do not yet know
		of?
		As partially discussed in Section~\ref{sec:generalization}, we first need to define to which family
		of instances we want to generalize over.
		For instance, we might decide to learn a cutting plane selection policy for Euclidian \gls{tsp}
		problems.
		Even so, it remains a complex effort to generate problems that capture the essence of real
		applications.
		Moreover, \gls{co} problems are high dimensional, highly structured, and troublesome to visualize.
		The sole exercise of generating graphs is already a complicated one!
		The topic has nonetheless received some interest.
		\cite{SmithGenerating2015} claim that the confidence we can put in an algorithm
		``\textit{depends on how carefully we select test instances}'', but note however that too often, a
		new algorithm is claimed ``\textit{to be superior by showing that it outperforms previous approaches
		on a set of well-studied instances}''.
		The authors propose a problem instance generating method that consists of: defining an instance
		feature space, visualizing it in two dimensions (using dimensionality reduction techniques such as
		principal component analysis), and using an evolutionary algorithm to drive the instance generation
		toward a pre-defined sub-space.
		The authors argue that the method is successful if the easy and hard instances can be easily
		separated in the reduced instance space.
		The methodology is then fruitfully applied to graph-based problems, but would require redefining
		evolution primitives in order to be applied to other type of problems.
		On the contrary, \cite{MalitskyStructurePreserving2016} propose a method to generate problem
		instances from the same probability distribution, in that case, the one of ``\textit{industrial}''
		boolean satisfiability problem instances.
		The authors use a large neighborhood search, using destruction and reparation primitives, to search
		for new instances.
		Some instance features are computed to classify whether the new instances fall under the same cluster as
		the target one.

		Deciding how to represent the data is also not an easy task, but can have a dramatic impact on
		learning.
		For instance, how does one properly represent a \gls{bnb} node, or even the whole \gls{bnb} tree?
		These representations need to be expressive enough for learning, but at the same time, concise
		enough to be used frequently without excessive computations.

\section{Conclusions}
	\label{sec:conclusions}
	We have surveyed and highlighted how machine learning can be used to build \acrlong{co} algorithms
	that are partially learned.
	We have suggested that imitation learning alone can be valuable if the policy learned is
	significantly faster to compute than the original one provided by an expert, in this case a
	\acrlong{co} algorithm.
	On the contrary, models trained with a reward signal have the potential to outperform current
	policies, given enough training and a supervised initialization.
	Training a policy that generalizes to unseen problems is a challenge, this is why we believe
	learning should occur on a distribution small enough that the policy could fully exploit the
	structure of the problem and give better results.
	We believe end-to-end \acrlong{ml} approaches to \acrlong{co} can be improved by using
	machine learning in combination with current \acrlong{co} algorithms to benefit from the
	theoretical guarantees and state-of-the-art algorithms already available.

	Other than performance incentives, there is also interest in using machine learning as a modelling
	tool for discrete optimization, as done by \cite{LombardiBoostingCombinatorialProblem2018}, or to
	extract intuition and knowledge about algorithms as mentioned in
	\cite{BonamiLearningClassificationMixedInteger2018, KhalilLearningCombinatorialOptimization2017}.

	Although most of the approaches we discussed in this paper are still at an exploratory level of
	deployment, at least in terms of their use in general-purpose (commercial) solvers, we strongly
	believe that this is just the beginning of a new era for combinatorial optimization algorithms.

\section*{Acknowledgments}
	The authors are grateful to Emma Frejinger, Simon Lacoste-Julien, Jason Jo, Laurent Charlin, Matteo
	Fischetti, Rémi Leblond, Michela Milano, Sébastien Lachapelle, Eric Larsen, Pierre Bonami, Martina
	Fischetti, Elias Khalil, Bistra Dilkina, Sebastian Pokutta, Marco Lübbecke, Andrea Tramontani,
	Dimitris Bertsimas and the entire CERC team for endless discussions on the subject and for reading
	and commenting a preliminary version of the paper.

\bibliography{ref.bib}

\end{document}